\documentclass[letterpaper,journal]{IEEEtran}
\usepackage{amsmath,amsfonts}
\usepackage{algorithmic}
\usepackage{algorithm}
\usepackage{array}
\usepackage{amssymb}
\usepackage{bm}
\usepackage{booktabs} 
\usepackage{cite}
\usepackage{color}
\usepackage{graphicx}
\usepackage[colorlinks,citecolor=blue,linkcolor=blue,urlcolor=blue]{hyperref}
\usepackage{multirow}
\usepackage[caption=false,font=scriptsize,labelfont=sf,textfont=sf]{subfig}
\usepackage{stfloats}
\usepackage{textcomp}
\usepackage{tabularx}
\usepackage{url}
\usepackage{verbatim}
\usepackage{makecell}
\usepackage{orcidlink}

\begin{document}

\title{MapSR: Prompt-Driven Land Cover Map Super-Resolution via Vision Foundation Models}

\author{Ruiqi Wang\,\orcidlink{0009-0001-0488-7839}, Qi Yu, Jie Ma\,\orcidlink{0000-0001-7957-4274},~\IEEEmembership{Member,~IEEE}, Hanlin Wu\,\orcidlink{0000-0002-3505-0521},~\IEEEmembership{Member,~IEEE}    
    \thanks{This work was supported in part by the National Natural Science Foundation of China under Grant 62401064, and in part by the Fundamental Research Funds for the Central Universities under Grant 2024JJ040 and Grant 2024TD001. \emph{(Corresponding author: Hanlin Wu.)}
    }
    \thanks{The authors are with the School of Information Science and Technology, Beijing Foreign Studies University, Beijing 100089, China (e-mail: hlwu@bfsu.edu.cn).}
}

\markboth{}%
{}

\IEEEpubid{}

\maketitle

\begin{abstract}
    High-resolution (HR) land-cover mapping is often constrained by the high cost of dense HR annotations. We revisit this problem from the perspective of \textbf{map super-resolution}, which enhances coarse low-resolution (LR) land-cover products into HR maps at the resolution of the input imagery. Existing weakly supervised methods can leverage LR labels, but they typically use them to retrain dense predictors with substantial computational cost. We propose \textbf{MapSR}, a prompt-driven framework that decouples supervision from model training. MapSR uses LR labels once to extract class prompts from frozen vision foundation model features through a lightweight linear probe, after which HR mapping proceeds via training-free metric inference and graph-based prediction refinement. Specifically, class prompts are estimated by aggregating high-confidence HR features identified by the linear probe, and HR predictions are obtained by cosine-similarity matching followed by graph-based propagation for spatial refinement. Experiments on the Chesapeake Bay dataset show that MapSR achieves \textbf{59.64\% mIoU} without any HR labels, remaining competitive with the strongest weakly supervised baseline and surpassing a fully supervised baseline. Notably, MapSR reduces trainable parameters by four orders of magnitude and shortens training time from hours to minutes, enabling scalable HR mapping under limited annotation and compute budgets. The code is available at \url{https://github.com/rikirikirikiriki/MapSR}.
\end{abstract}


\section{Introduction}
\IEEEPARstart{W}{ith} the increasing spatial resolution of remote-sensing imagery, scalable high-resolution (HR) land-cover mapping has become increasingly important for Earth surface monitoring and related applications \cite{zhang2013analysis}.

Obtaining dense HR annotations at scale remains costly and labor-intensive. Despite this cost, most existing HR mapping methods still formulate land-cover mapping as a dense prediction problem and directly train supervised models on HR labels, ranging from early pixel-wise classifiers such as random forests \cite{chan2008evaluation} to recent CNN- and Transformer-based architectures \cite{robinson2019large,wang2022novel}. While effective, these methods depend heavily on large-scale HR annotation and are difficult to scale to new regions.

\begin{figure}[t]
    \centering
    \includegraphics[width=0.9\linewidth]{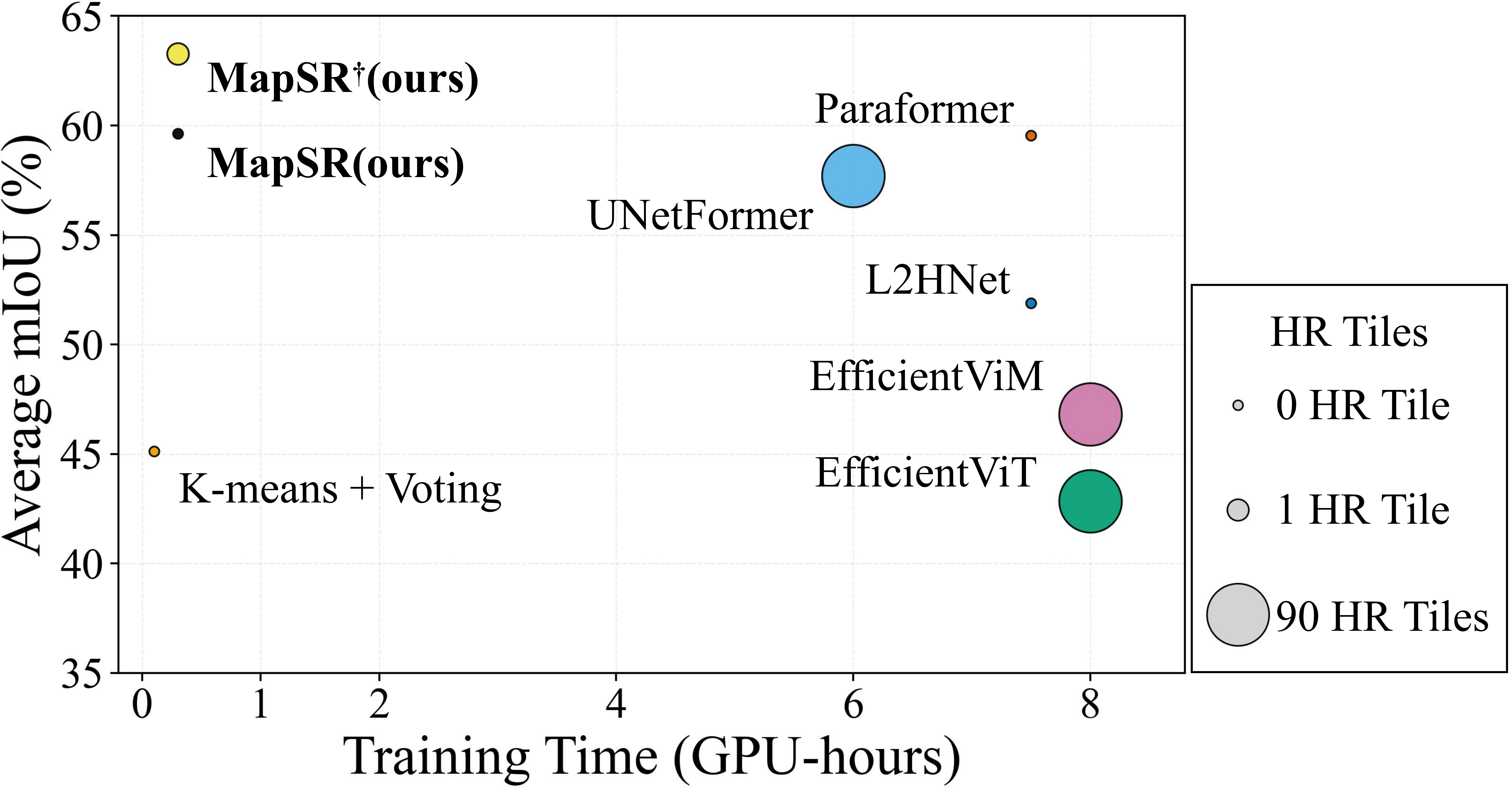}
    \caption{Comparison of representative methods in terms of annotation cost and training time for HR land-cover mapping, highlighting the efficiency advantage of MapSR.}
    \label{fig:compare}
\end{figure}

To alleviate this annotation burden, recent methods exploit coarse LR land-cover products to improve HR mapping accuracy \cite{Paraformer,L2HNet}. However, they are still typically implemented within weakly supervised dense prediction frameworks, where LR supervision is used to retrain large segmentation models (on the order of $10^{7}$ parameters) through gradient descent, typically requiring several GPU-hours. As a result, these methods reduce annotation cost without substantially reducing training cost. In other words, even when coarse LR supervision is available, the cost of retraining a large dense predictor remains high. This observation suggests that the main bottleneck is not only the lack of HR labels, but also the prevailing train-then-infer paradigm itself. Rather than treating LR products primarily as supervision for optimizing a dense predictor, we revisit the problem from the perspective of \emph{map super-resolution}, where coarse LR products are treated as the source to be enhanced into HR land-cover maps.

Recent advances in vision foundation models, especially self-supervised vision transformers (ViTs)~\cite{dosovitskiy2020image} such as DINOv2 \cite{oquab2023dinov2}, have shown that frozen features can provide strong semantic transfer across downstream tasks. Recent work further suggests that upsampled DINOv2 features can already support unsupervised vision tasks and weakly supervised segmentation without retraining a full model \cite{caron2021emerging, docherty2025upsampling}. This suggests that LR labels may only be needed once to identify class semantics in the frozen embedding space, turning map super-resolution into a lightweight prompt construction problem rather than a heavy weakly supervised learning problem.

Based on this idea, we propose MapSR, a prompt-driven framework for HR land-cover map super-resolution via vision foundation models. MapSR freezes the feature extractor, uses a lightweight linear probe (only 4K trainable parameters) to identify high-confidence HR pixels under LR supervision, aggregates them into class prompts, and performs HR prediction by cosine-similarity matching followed by graph-based refinement. In this way, supervision is distilled into compact class prototypes rather than full model weights, so new images can be processed without gradient-based optimization (see Fig.~\ref{fig:compare}). The main contributions are as follows:
\begin{enumerate}
    \item We revisit HR land-cover mapping from the perspective of \emph{map super-resolution} and formulate it as enhancing coarse LR land-cover products into HR maps, rather than treating them solely as supervision for dense prediction.
    \item We propose a prompt-driven paradigm that decouples supervision from model training: LR labels are used once to extract class prompts through a lightweight linear probe, enabling subsequent HR mapping via metric-based inference rather than full-network retraining.
    \item We develop a sequential framework that combines frozen vision foundation model features, attention-based feature upsampling, prompt-based pixel classification, and graph-based refinement to produce spatially coherent HR land-cover maps.
    \item Extensive experiments demonstrate that this paradigm achieves competitive accuracy without HR labels while reducing trainable parameters to only 4K and training time to about 18 minutes.
\end{enumerate}

\section{Methodology}

\subsection{Problem Formulation}
Given an HR image $I\in\mathbb{R}^{3\times H\times W}$ and a coarse LR land-cover product $Y_\text{LR}\in\mathcal{Y}^{h\times w}$, where $h\ll H$ and $w\ll W$ (typically by a factor of 10--30), the goal of \emph{map super-resolution} is to recover an HR land-cover map $Y\in\mathcal{Y}^{H\times W}$ by enhancing the LR product to the resolution of the input image. Here, $\mathcal{Y}=\{0,1,\ldots,n\}$ denotes the discrete label space of $C=n+1$ land-cover classes.

The key distinction of our formulation lies in how the LR product is used. Instead of treating it as supervision for retraining a dense predictor, we use it once to extract class prompts, after which HR mapping is performed by metric-based inference in the frozen embedding space.

\begin{figure*}[t]
    \centering
    \includegraphics[width=1\textwidth]{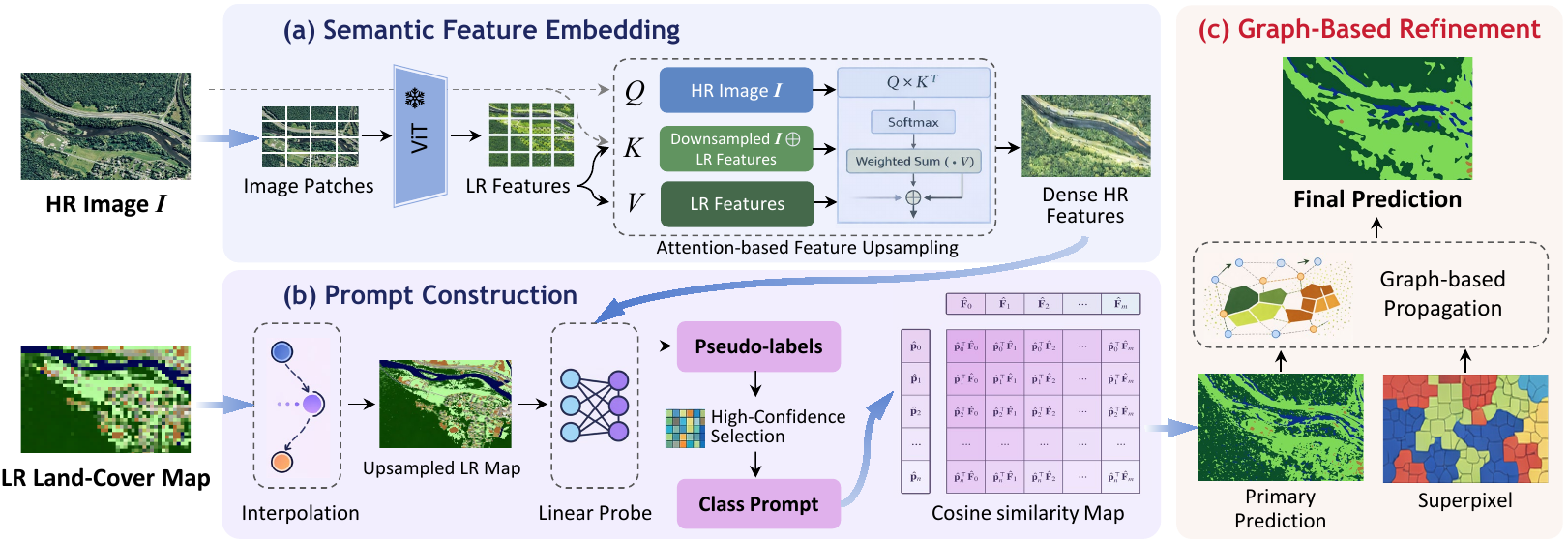}
    \caption{Overview of MapSR. (a) Feature extraction and upsampling produce dense HR features from the input image using a frozen vision foundation model. (b) Prompt construction uses the coarse LR land-cover product to extract class prompts for cosine-similarity-based initial prediction. (c) Graph-based prediction refinement propagates prediction scores over a superpixel graph to obtain the final HR land-cover map.}
    \label{fig:framework}
\end{figure*}

To this end, we extract a dense HR feature map $\bm{F}\in\mathbb{R}^{D\times H\times W}$ from the input image using a frozen ViT encoder and a feature upsampling module. As illustrated in Fig.\,\ref{fig:framework}, the proposed approach consists of three steps: (1) constructing class prompts $\{\hat{\bm{p}}_c\}_{c=0}^n$ from the LR product via lightweight linear probing; (2) classifying HR pixels via nearest-neighbor search against prompts; and (3) refining the initial predictions through graph-based spatial regularization. Formally, the mapping is expressed as
\begin{equation}
    f(I, Y_\text{LR}) = g\left(\arg\max_{c} \langle \hat{\bm{F}}(I), \hat{\bm{p}}_c(Y_\text{LR}) \rangle\right),
\end{equation}
where $\hat{\bm{F}}$ and $\hat{\bm{p}}_c(Y_\text{LR})$ denote the $\ell_2$-normalized HR feature map and the class prompt extracted from the LR product, respectively, and $g(\cdot)$ is the graph-based refinement operator.

\subsection{Semantic Feature Embedding}

Self-supervised ViTs like DINOv2~\cite{oquab2023dinov2} provide robust, semantically rich features, but their patch-level outputs are too coarse for HR land-cover mapping. We therefore upsample them into dense, spatially aligned features using the HR image structure.

\subsubsection{Foundation Model as Feature Extractor}
We adopt a frozen ViT encoder $\phi_\text{ViT}$ pre-trained via self-supervised learning \cite{oquab2023dinov2}. Given an input image $I$, the encoder produces patch-level tokens $\{t_i\}_{i=1}^{h_p w_p}$ where each $t_i \in \mathbb{R}^D$ corresponds to a non-overlapping patch. After discarding the CLS token and reshaping, we obtain:
\begin{equation}
    \bm{F}_\text{LR} = \operatorname{Reshape}(\phi_\text{ViT}(I)) \in \mathbb{R}^{D \times h_p \times w_p},
\end{equation}
where $h_p = H/p$ and $w_p = W/p$ for patch size $p$.

\subsubsection{Attention-based Feature Upsampling}

The patch-level features $\bm{F}_\text{LR}$ are much coarser than the target resolution $H \times W$. Rather than using bilinear interpolation, we employ an attention-based feature upsampling module \cite{wimmer2025anyup} that leverages the HR image $I$ as guidance to densify $\bm{F}_\text{LR}$ in a content-aware manner.

Specifically, pixel-wise query features are extracted from $I$, while keys are derived from a downsampled version of $I$ fused with $\bm{F}_\text{LR}$. The values are taken directly from $\bm{F}_\text{LR}$ without modification. The resulting dense feature map $\bm{F} \in \mathbb{R}^{D \times H \times W}$ is then computed via cross-attention:
\begin{equation}
    \bm{F}_{:,u,v} = \sum_{i,j} \alpha_{(u,v)\rightarrow(i,j)} \cdot \bm{F}_{\text{LR},:,i,j},
\end{equation}
where the attention weights $\alpha_{(u,v)\rightarrow(i,j)}$ are obtained from the similarity between the HR query at $(u,v)$ and the LR key at $(i,j)$. This mechanism ensures that $\bm{F}$ preserves the semantic content of $\bm{F}_\text{LR}$ while remaining spatially aligned with the HR image structure.

\subsection{Class Prompt Construction from LR Labels}
\label{sec:class_prompt_construction}

The central challenge in map super-resolution is to transfer semantic information from a coarse LR product to an HR land-cover map. We address this by constructing class prompts—representative embeddings for each land-cover class—that capture class semantics from the LR product while remaining applicable to HR pixel classification.

From a metric learning perspective, each class prompt can be viewed as a prototype in the embedding space. Under $\ell_2$ distance, the ideal prompt for class $c$ is the mean embedding of HR pixels belonging to that class:
\begin{equation}
    \bm{p}_c^* = \mathbb{E}_{I, Y^*} \left[ \bm{F}_{:,u,v} \big| Y_{u,v}^* = c \right]
    = \int_{\mathcal{Z}} \bm{z} p(\bm{z}|c) \mathrm{d}\bm{z},
\end{equation}
where $Y^*\in \mathcal{Y}^{H\times W}$ denotes the (unavailable) HR ground-truth land-cover map, $\bm{F}_{:,u,v}$ is the feature embedding of pixel $(u,v)$, and $p(\bm{z}|c)$ is the conditional probability density of embeddings given class $c$.

Because $Y^*$ is unavailable, $\bm{p}_c^*$ cannot be computed directly. We therefore estimate class prompts from the coarse LR product in two steps. First, LR supervision is used once to train a lightweight linear probe on frozen HR features. Second, the trained probe is applied to identify HR pixels whose predictions agree with the upsampled LR labels, and their features are aggregated to form class prompts. In this way, the LR product is used only for one-time prompt estimation rather than iterative model updating.

\subsubsection{Linear Probing with LR Labels}
We train a lightweight linear classifier $W \in \mathbb{R}^{C \times D}$ on the frozen HR features $\bm{F}$, using the LR product only as a one-time supervision source. After upsampling $Y_\text{LR}$ to the HR grid via nearest-neighbor interpolation, we denote the resulting noisy pixel-wise label at location $(u,v)$ as $\ell_{u,v} = Y_\text{LR}^\uparrow(u,v)$. The cross-entropy loss is then:
\begin{equation}
    \mathcal{L}(W) = -\sum_{(I,Y_\text{LR}) \in \mathcal{D}_\text{LR}}
    \sum_{u,v}
    \log \frac{\exp\big( W_{\,\ell_{u,v}}^{\top} \bm{F}_{:,u,v} \big)}
    {\sum_{c'=0}^{n} \exp\big( W_{c'}^{\top} \bm{F}_{:,u,v} \big)},
\end{equation}
where $\mathcal{D}_\text{LR}$ is the training dataset of LR-labeled images.

\subsubsection{High-Confidence Prompt Aggregation}
After training, we apply the linear probe $W$ to the full-resolution dense features $\bm{F}\in\mathbb{R}^{D\times H\times W}$ to obtain HR class predictions. For each class $c$, we retain only pixels whose linear-probe prediction agrees with the upsampled LR label:
\begin{equation}
    \Omega_c=\left\{(u,v)\mid \arg\max_{c'}W_{c'}^{\top}\bm{F}_{:,u,v}=c,\ \ell_{u,v}=c\right\}.
\end{equation}
The class prompt is then computed as the empirical mean of high-confidence features:
\begin{equation}
    \hat{\bm{p}}_c=\frac{1}{|\Omega_c|}\sum_{(u,v)\in\Omega_c}\bm{F}_{:,u,v}.
\end{equation}

Together, linear probing and high-confidence selection provide a simple denoising mechanism for prompt estimation. Since upsampled LR labels usually preserve the dominant class within each region, aggregating only high-confidence predictions yields class prompts that better approximate the underlying HR semantic distribution.

\subsection{Metric-Based Initial Prediction}

Given the upsampled features $\bm{F}$ and class prompts $\{\hat{\bm{p}}_c\}_{c=0}^n$, we perform an initial pixel-wise classification via nearest-class-center assignment in the embedding space. After $\ell_2$ normalization, cosine similarity becomes a natural measure of semantic affinity:
\begin{equation}
    s_c(u,v) = \frac{\hat{\bm{p}}_c^{\top}\bm{F}_{:,u,v}}{\|\hat{\bm{p}}_c\|_2\|\bm{F}_{:,u,v}\|_2} = \hat{\bm{p}}_c^{\top}\hat{\bm{F}}_{:,u,v},
\end{equation}
where $\hat{\bm{p}}_c$ and $\hat{\bm{F}}_{:,u,v}$ are $\ell_2$-normalized vectors. The initial prediction is:
\begin{equation}
    \hat{Y}_{u,v} = \arg\max_{c\in\{0,\ldots,n\}} s_c(u,v).
\end{equation}

This yields an initial land-cover map, which is subsequently refined through graph-based prediction refinement.

\subsection{Graph-Based Prediction Refinement}

Our initial HR predictions are based on local feature--prompt similarities and may therefore be unstable near class boundaries. To address this, we refine the prediction scores through graph-based propagation \cite{zhou2003learning,stojnic2025lposs}, which encourages nearby and semantically similar regions to produce consistent predictions while remaining close to the initial estimate.

\subsubsection{Graph Construction}
We construct an undirected graph $G=(V,E)$ where each node corresponds to a superpixel generated by SLIC \cite{achanta2012slic}, which reduces computational cost while encouraging local semantic homogeneity. Let $\{z_i\}_{i=1}^N$ and $\{x_i\}_{i=1}^N$ denote the mean feature embeddings and spatial coordinates of each superpixel. Edges are established via $k$-nearest neighbors in a joint feature-spatial space, with weights:
\begin{equation}
    A_{ij} = \underbrace{\big(\max(0, z_i^{\top}z_j)\big)^{\gamma}}_{\text{Feature similarity}} \cdot \underbrace{\exp\!\left(-\sigma\|x_i-x_j\|_2^{p}\right)}_{\text{Spatial proximity}}.
\end{equation}
Here, the parameter $\gamma>0$ sharpens high-similarity edges, and $\sigma, p > 0$ control the spatial decay rate.

\subsubsection{Prediction Propagation}
Let $\hat{Y}\in\mathbb{R}^{N\times C}$ denote the initial superpixel logits, obtained by averaging pixel-level prediction scores within each superpixel. We seek refined logits $\tilde{Y}\in\mathbb{R}^{N\times C}$ that balance fidelity to $\hat{Y}$ and smoothness over the graph \cite{zhou2003learning}:
\begin{equation}
    \tilde{Y} = \arg\min_{\tilde{Y}}\bigg(\|\tilde{Y}-\hat{Y}\|_F^2 + \lambda \sum_{(i,j)\in E} A_{ij} \|\tilde{Y}_i - \tilde{Y}_j\|_2^2\bigg).
\end{equation}
This objective admits a closed-form solution via the linear system:
\begin{equation}
    (\bm{I}_N - \alpha \hat{A}) \tilde{Y} = (1-\alpha)\hat{Y},
\end{equation}
where $\hat{A}=D^{-1/2}AD^{-1/2}$ is the symmetrically normalized adjacency matrix, $D_{ii}=\sum_j A_{ij}$ is the degree matrix, and $\alpha = \lambda / (1+\lambda)$. This can be efficiently solved by the fixed-point iteration:
\begin{equation}
    \tilde{Y}^{(t+1)} = \alpha \hat{A} \tilde{Y}^{(t)} + (1-\alpha)\hat{Y},
\end{equation}
which preserves the same balance between smoothness and fidelity.

After convergence, the refined prediction for each superpixel is obtained by taking the class with the highest logit. This graph-based refinement effectively suppresses noise and enforces spatial consistency, particularly along class boundaries where the initial predictions are most uncertain.

\begin{table*}[t]
    \centering
    \caption{Performance Comparison on Trainable Params (M) and Six Regional Subsets of the Chesapeake Bay Dataset (mIoU, \%)}
    \setlength{\tabcolsep}{5.5pt}
    \begin{tabular}{cccccccccc}
        \toprule
        \textbf{Category} & \textbf{Method} & \textbf{\makecell{Trainable\\Params (M)}} & \textbf{Delaware} & \textbf{New York} & \textbf{Maryland} & \textbf{Pennsylvania} & \textbf{Virginia} & \textbf{West Virginia} & \textbf{Average}\\
        \midrule
        \multirow{1}{*}{Unsupervised}
        & K-means\,+\,voting & 0 & 32.70 & 50.87 & 46.74 & 51.34 & 43.45& 45.53 & 45.11\\
        \midrule
        \multirow{2}{*}{\makecell{Weakly-\\Supervised}}
        & Paraformer\cite{Paraformer} & 145.4 & 54.95 & 63.83 & 66.10 & 56.83 & \textbf{66.23} & 49.27 & 59.54\\
        & L2HNet\cite{L2HNet} & 36.92 & 46.37 & 59.87 & 44.16 & 49.00 & \underline{65.09}& 46.80 & 51.88\\
        \midrule
        \multirow{3}{*}{\makecell{Fully-\\Supervised}}
        & EfficientViT\cite{cai2023efficientvit} & 51.31 & 48.13 & 40.65 & 45.48 & 45.04 & 45.66 & 32.10 & 42.84\\
        & EfficientViM\cite{lee2025efficientvim} & 21.70 & 46.81 & 53.43 & 50.41 & 44.28 & 45.09 & 40.87 & 46.82\\
        & UNetFormer\cite{wang2022unetformer} & 11.72 & 55.08 & \underline{66.88} & \underline{66.83} & 60.04 & 54.70 & 42.66 & 57.70\\
        \midrule
        \multirow{2}{*}{\makecell{Prompt-\\Driven}}
        & MapSR & {0.004} & \underline{60.91} & 62.40 & 66.19 & \underline{60.26} & 57.78 & \underline{50.31} & \underline{59.64}\\
        & MapSR$^\dagger$ & 0 & \textbf{63.51} & \textbf{67.91} & \textbf{70.19} & \textbf{65.78} & 57.92 & \textbf{54.31} & \textbf{63.27}\\
        \bottomrule
    \end{tabular}
    \label{tab:ours_vs_baselines}
    \vspace{-1em}
\end{table*}

\section{EXPERIMENTS AND ANALYSIS}
\subsection{Datasets}
Our experiments are conducted on the Chesapeake Bay dataset~\cite{claggett2025chesapeake}, which covers the Chesapeake Bay watershed and surrounding areas in the United States. The dataset is divided into six regional subsets corresponding to six states in the study area and contains 732 large, non-overlapping tiles. Each tile is approximately $6000\times 7500$ pixels at 1~m spatial resolution. The dataset provides 1 m HR images from NAIP, 30 m LR land-cover products from NLCD with 16 original classes, and 1 m ground-truth land-cover maps from the CCLC project.

\subsection{Experimental Setup}
In our MapSR framework, we adopt DINOv2 with a feature dimension of 768 as the frozen ViT backbone for feature extraction \cite{oquab2023dinov2}. Following Paraformer~\cite{Paraformer}, the original 16 NLCD/CCLC categories are merged into 5 evaluation classes for all compared methods. A sliding-window strategy with a chip size of 448 is adopted during graph-based prediction refinement to avoid excessive memory consumption, and the outputs are mosaicked to obtain the final result. For graph-based refinement, SLIC superpixels \cite{achanta2012slic} are generated with $n_{\text{segments}}=8000$ and $\text{compactness}=10$. A $k$-NN graph is constructed with $k=100$, and prediction propagation is performed with $\alpha=0.5$ (i.e., $\lambda=1$) and $\gamma=1.0$. For the spatial kernel, we fix $p=2$ and set $\sigma=1$ after coordinate normalization.

For each of the six subregions in the Chesapeake Bay dataset, we randomly select ten HR land-cover maps as the test set. Fully supervised baselines (EfficientViT~\cite{cai2023efficientvit}, EfficientViM~\cite{lee2025efficientvim}, and UNetFormer~\cite{wang2022unetformer}) are trained on the remaining HR CCLC labels (excluding the test tiles). Weakly supervised baselines (Paraformer~\cite{Paraformer} and L2HNet~\cite{L2HNet}) are trained using all available LR NLCD labels as supervision, without accessing HR ground truth. In contrast, MapSR uses the remaining LR products only for prompt construction rather than full-model training, and an additional variant, MapSR$^\dagger$, uses one non-test HR tile to directly estimate class prompts by averaging the features of labeled HR pixels, providing an empirical upper bound under minimal HR supervision. Since no existing unsupervised methods are available for this setting, we construct an unsupervised baseline by applying K-means clustering \cite{kmeans} and assigning each cluster to a class via majority voting from the corresponding LR labels.

\subsection{Comparison Results}

Here, the trainable parameter count for MapSR refers to the lightweight linear probe only ($C\times D$), while class prompts are computed by feature averaging without optimization. Table~\ref{tab:ours_vs_baselines} summarizes the quantitative comparison. MapSR reaches 59.64\% mIoU without any HR labels, remaining competitive with the strongest weakly supervised baseline (Paraformer, 59.54\%) and outperforming a fully supervised baseline (UNetFormer, 57.70\%). With only \emph{one} non-test HR tile for prompt estimation, MapSR$^\dagger$ further improves to 63.27\% (+3.63\%), serving as an empirical upper bound under minimal HR supervision. Notably, MapSR optimizes only $\sim 10^{3}$ parameters, and the lightweight linear probe training takes about 18 minutes on a single RTX 4090 GPU, while deep segmentation baselines typically require $\sim 10^{7}$ parameters and 6--8 GPU-hours, making our paradigm highly practical for large-area HR mapping under limited annotation and compute budgets.

\subsection{Ablation Study}
\begin{figure}[t]
    \centering
    \includegraphics[width=1\linewidth]{figures/fig3.png}
    \caption{Visual comparison for HR land-cover mapping. (a) HR image; (b) HR ground-truth label; (c) K-means\,+\,voting; (d) EfficientViT; (e) Paraformer; (f) UNetFormer; (g) MapSR; (h) MapSR without graph-based refinement.}
    \label{fig:ablation}
\end{figure}

\begin{table}[t]
    \centering
    \caption{Ablation of feature upsampling, graph refinement, and superpixels on Chesapeake Bay (New York)}
    \setlength{\tabcolsep}{8pt}
    \label{tab:ablation_ny_chesapeake}
    \begin{tabular}{ccc|c}
        \toprule
        \textbf{Feature Upsampling} & \textbf{Graph Refine} & \textbf{Superpixel} & \textbf{mIoU (\%)} \\
        \midrule
        &  &  & 57.70 \\
        \checkmark &  &  & 61.77 \\
        \checkmark & \checkmark &  & 62.26 \\
        \checkmark & \checkmark & \checkmark & \textbf{62.40} \\
        \bottomrule
    \end{tabular}
\end{table}

To evaluate the contribution of each component, we perform ablation experiments on the New York subset of the Chesapeake Bay dataset (Table~\ref{tab:ablation_ny_chesapeake}).

First, we investigate the impact of feature upsampling. As a reference, a prompt-based variant without feature upsampling attains 57.70\% mIoU. With feature upsampling and without graph-based prediction refinement, MapSR reaches 61.77\%, indicating that precise feature alignment is fundamental for HR mapping.

Second, graph-based prediction refinement further improves the average mIoU from 61.77\% to 62.26\%, and incorporating superpixels improves it again to 62.40\%. As shown in Fig.~\ref{fig:ablation}, the raw MapSR prediction contains isolated noise and irregular boundaries. After graph-based refinement, the results become cleaner and more spatially consistent, demonstrating its capability to suppress local noise and enhance boundary delineation.

\section{Conclusion}
This paper proposes MapSR, a prompt-driven framework for HR land-cover mapping that reformulates the task as map super-resolution over coarse LR products. The key idea is to decouple supervision from model training: LR labels are used once to extract class prompts through a lightweight linear probe, after which HR mapping proceeds via training-free metric inference and graph-based refinement. Experiments on the Chesapeake Bay dataset show that MapSR remains competitive with fully and weakly supervised methods while substantially reducing training cost, enabling scalable HR land-cover mapping without HR annotations and under limited compute budgets. Future work will explore more robust prompt construction under temporal mismatch and broader cross-region generalization.

\bibliographystyle{IEEEtran}
\bibliography{refs}

\end{document}